\pdfoutput=1

\documentclass[11pt]{article}

\usepackage{EACL2023}

\usepackage{times}
\usepackage{latexsym}
\usepackage{booktabs}
\usepackage{amssymb}

\usepackage[T1]{fontenc}

\usepackage[utf8]{inputenc}

\usepackage{microtype}

\usepackage{inconsolata}

\usepackage[pdftex]{graphicx}

\usepackage {xcolor}

\usepackage{tabularx}
\usepackage{multicol}
\usepackage{multirow}
\usepackage{longtable}

\usepackage{amsmath}

\usepackage{tipa}

\usepackage{adjustbox}

\title{Information-Theoretic Characterization of Vowel Harmony: \\ A Cross-Linguistic Study on Word Lists}

\author{Julius Steuer\textsuperscript{\textipa{U}} \hspace{0.6cm} Badr M. Abdullah\textsuperscript{\textipa{U}}  \hspace{0.6cm} Johann-Mattis List\textsuperscript{\textipa{Y}} \hspace{0.6cm} Dietrich Klakow\textsuperscript{\textipa{U}} \\      
\textbf{\textsuperscript{\textipa{U}}}Language Science and Technology (LST), Saarland University \\   \textbf{\textsuperscript{\textipa{Y}}}MPI-EVA,  Univ. of Passau \\ %
       \normalsize{\textsf{\{ jsteuer,  babdullah, dietrich \}@lsv.uni-saarland.de}, \textsf{mattis.list@uni-passau.de}} 
}

\begin{document}
\maketitle
\begin{abstract}
We present a cross-linguistic study that aims to quantify vowel harmony  using data-driven computational modeling.
Concretely, we define an information-theoretic measure of harmonicity based on the predictability of vowels in a natural language lexicon, which we estimate using phoneme-level language models (PLMs). 
Prior quantitative studies have relied heavily on inflected word-forms in the analysis of vowel harmony. 
We instead train our models using cross-linguistically comparable lemma forms with little or no inflection, which enables us to cover more under-studied languages. 
Training data for our PLMs consists of word lists with a maximum of 1000 entries per language. 
Despite the fact that the data we employ are substantially smaller than previously used corpora, our experiments demonstrate the neural PLMs capture vowel harmony patterns in a set of languages that exhibit this phenomenon. 
Our work also demonstrates that word lists are a valuable resource for typological research, and offers new possibilities for future studies on low-resource, under-studied languages.

\end{abstract}

\section{Introduction}

\subsection{Vowel Harmony}
Many of the world's languages exhibit vowel harmony -- a phonological co-occurrence constraint whereby vowels in polysyllabic words have to be members of the same natural class \cite{ohala_towards_1994}. 
Natural classes of vowels are defined with respect to polar phonological features such as vowel backness (\textsc{$\pm$back}) and roundedness (\textsc{$\pm$round}). 
In a prototypical language with backness, or $\pm\textsc{back}$ harmony, all vowels within a word tend to share the $\pm\textsc{back}$ feature, i.e. they are either all front ($-\textsc{back}$) or back ($+\textsc{back}$).
Table \ref{tab:vh_turkish} illustrates vowel harmony in Turkish, one of the languages best known to have this feature. 
In Table \ref{tab:vh_turkish}, the nominative plural and genitive plural are examples of $-\textsc{back}$ harmony, while the genitive singular column of $+\textsc{back}$ harmony. In the case of Turkish, vowel harmony can be defined as a constraint applying to almost all words and the entire inflectional system. In other languages vowel harmony may be restricted to the inflectional system, or even only a subset of inflectional suffixes. 
For example, In Estonian there are vestiges of vowel harmony in lexical items and it is absent from the inflectional system, while in Bislama it only occurs in a single suffix marking transivity \citep{crowley_bislama_2017}. Between these extremes of Turkish and Bislama lie languages such as Finnish and Hungarian, with intermediate vowel harmony systems where not all vowels participate in vowel harmony to the same extent. Both languages have $\pm\textsc{back}$ harmony, but a subset of the $-\textsc{back}$ vowels allow $+\textsc{back}$ harmony to spread: In a word like [l\textipa{A}tik\textipa{:}o] `box' (not [l\textipa{A}tik\textipa{:}ø]), $+\textsc{back}$ harmony is not violated, whereas a word containing only neutral vowels triggers $-\textsc{back}$ harmony, as in [merkitys] `meaning' where the $+\textsc{\text{back}}$ disharmonic form [merkitus] is not possible.

\begin{table*}[!ht]
    \centering
    \begin{tabular}{lccccc}
        \hline
        & \textbf{Nom. Sg.} & \textbf{Gen. Sg.} & \textbf{Nom. Pl.} & \textbf{Gen. Pl.} & \textbf{Gloss} \\
        \hline\hline 
         $-\textsc{back}$/$-\textsc{round}$ & {[ip]}  &  [ip-in]   &  [ip-l\textipa{E}r]   &  [ip-l\textipa{E}r-in]  & 'string'  \\
         $+\textsc{back}$/$-\textsc{round}$ & {[k\textipa{W}z]} &  [k\textipa{W}z-\textipa{W}n]  &  [k\textipa{W}z-lar]  &  [k\textipa{W}z-lar-\textipa{W}n] & 'girl'  \\
         $-\textsc{back}$/$+\textsc{round}$ & {[jyz]} &  [jyz-yn]  &  [jyz-l\textipa{E}r]  &  [jyz-l\textipa{E}r-in] & 'face' \\
         $+\textsc{back}$/$+\textsc{round}$ & {[pul]} &  [pul-un]  &  [pul-lar]  &  [pul-lar-\textipa{W}n] & 'stamp' \\
        \hline
    \end{tabular}
    \caption{Illustration of the Turkish vowel harmony system following \citet{polgardi_vowel_1999}. The first vowel of a word form determines the harmony type. If the first vowel is $+\textsc{back}$, the vowels of the following suffixes must agree w. r. t. the $+\textsc{back}$ feature. $\pm\textsc{round}$ harmony applies only in suffixes that have separate forms for this feature: The genitive suffix takes both $\pm\textsc{back}$ and $\pm\textsc{round}$ forms, while the plural suffix varies only for $\pm\textsc{back}$.}
    \label{tab:vh_turkish}
\end{table*}

The rather broad application of the term has made it increasingly difficult to define it as a phonological process (cf. \citealt{anderson_problems_1980_new}). If vowel harmony is used as a typological feature to group languages into phylogenetic families, this broad application becomes perilous to the researcher since they have to be aware of the the degree of vowel harmonicity in the individual languages. Instead of searching for a necessarily complex definition of vowel harmony, research has consequentially concentrated on a quantitative description.

\subsection{Prior Work and Scope}

Prior approaches to a quantitative description of vowel harmony have mostly focused on strictly local harmony processes. \citet{Mayer2010Visua-12872} used vowel succession counts derived from corpora of inflected word-forms to quantify vowel harmony in a large number of languages in terms of $\chi^2$-values, while \citet{ozburn_segment-specific_2019} used count data to estimate succession probabilities and calculate the relative risk of encountering an harmonic vowel in a word form. These two approaches treated all positions in a word form identically. \citet{goldsmith_information_2012} argued that vowel harmony involves at least one type of non-local dependency, since it operates over consonants intervening between adjacent vowels. They employed a simple \textit{n}-gram language model to learn the phonology of Finnish and calculated pointwise mutual information of vowel-vowel and consonant-vowel pairs based on the phoneme probabilities predicted by the language model, finding evidence for consonant-vowel harmony besides the expected $\pm\textsc{back}$ harmony, with a small bias towards $+\textsc{back}$ harmony.
However, \textit{n}-gram language models are limited by their predefined context size. A language model with a left-hand context of $n=3$ cannot capture the effect of vowel harmony if it operates over a neutral vowel intervening between two harmonic vowels. While this effect could be mitigated by allowing by allowing for a larger or flexible $n$, estimating probabilities from corpora becomes increasingly difficult with higher values of $n$. 
In this study we aim to improve over these methods by quantifying vowel harmony with a information-theoretic measure based on \emph{surprisal}, capturing the relative strength of vowel harmony in language in terms of the likelihood of a vowel in a word to share a specific feature with preceding vowels. To do so, we employ neural recurrent language models with variable-length preceding phoneme context that are trained on cross-linguistically comparable lexical data.
While some previous work on modeling vowel harmony with language models has been carried out \citep{rodd_recurrent_1997}, 
finding evidence for Turkish vowel harmony in the hidden activations of a simple neural language model, it seems that this topic has not been further explored since then. 
In the following section, we first introduce feature surprisal as an information-theoretic measure of vowel harmony (\S{\ref{quan}}). 
We then present our computational experiments with the introduced measure of vowel harmony and discuss the results of their application to a large collection of cross-linguistic lexical data (\S{\ref{experi}}, \S{\ref{results}).
We conclude by discussing the implications of our study for future studies on vowel harmony in classical and computational studies (\S{\ref{conclu}}).

\section{Quantifying Vowel Harmony}
\label{quan}

\subsection{Phoneme-Level Language Models}
\textbf{Preliminaries and Notations.} \hspace{0.5cm} To quantify vowel harmony in our study, we make use of phoneme-level language models (PLMs). 
Consider a natural language with a lexicon $\mathcal{L}$ and a phoneme inventory $\boldsymbol{\Phi}$ (using IPA symbols). 
Using a cross-linguistic word list, we obtain $K$ samples from the lexicon  $\mathcal{D} = \{\boldsymbol{w}^k\}_{k=1}^{K} \sim \mathcal{L}$ where each sample is a word-form that is transcribed as a phoneme sequence $\boldsymbol{w} = (\varphi_1, \cdots, \varphi_{|\boldsymbol{w}|}) \in \boldsymbol{\Phi}^*$.
Given this sample of word-forms as training data, a PLM can be trained to estimate a probability distribution over $\boldsymbol{\Phi}$ by maximizing the term 
\begin{equation} \label{eq1}
\begin{split}
J(\theta, \mathcal{D}) & =  \sum_{\boldsymbol{w} \in \mathcal{D}} p(\boldsymbol{w}; \theta)  \\
 & = \sum_{\boldsymbol{w} \in \mathcal{D}} \quad \prod_{t \in \{1, \cdots, |\boldsymbol{w}|\}}  p(\varphi_t | \boldsymbol{\varphi}_{<t}; \theta)
\end{split}
\end{equation}
Here, $\theta$ are the parameters of the model that are learned by maximizing the objective function above. 
Once a PLM has been trained, it can be used to compute the probability of unseen, held-out word-forms (i.e, word-forms that were not observed in the training data).
Ideally, a PLM should assign a higher probability mass to plausible word-forms given the phonotactic rules of the language of the train data, and lower probability to implausible word-forms.

\vspace{0.3cm}
\noindent
\textbf{Recurrent PLMs.} \hspace{0.5cm}
Although different architectures can be used to build a PLM, we choose to employ a recurrent architecture based on unidirectional long short-term memory (LSTM) cell \cite{hochreiter1997long}. 
Given a word-form as a sequence of phonemes $\boldsymbol{w} = (\varphi_1, \cdots, \varphi_{|\boldsymbol{w}|})$, each phoneme is first projected into a continuous-vector phoneme representation using an embedding matrix as  $\mathbf{E}(\varphi_t) = \mathbf{x}_t \in \mathbb{R}^d$. 
Then, the LSTM takes as input the sequence at each position $t$ within the word-form to compute the hidden state representation 
\begin{equation}
    \mathbf{h}_t = \mathcal{F}_{\textsc{lstm}}(\mathbf{x}_t, \mathbf{h}_{t-1} ) \in \mathbb{R}^h
\end{equation}
To obtain a probability distribution over the phoneme inventory, a linear transformation is applied on the hidden state vector followed by a softmax function to obtain a probability vector as 

\begin{equation}
    p(\varphi_t | \boldsymbol{\varphi}_{<t}) = \textsc{softmax}(\mathbf{W}\mathbf{h}_t + \mathbf{b})
\end{equation}
Here, $\mathbf{W} \in \mathbb{R}^{|\mathbf{\Phi}| \times h}$ is a projection matrix at the network output and $\mathbf{b} \in \mathbb{R}^{|\mathbf{\Phi}|}$ is a bias term.

Nevertheless, we make a few  (trivial) design modifications to the vanilla LSTM-based PLMs to make them more suitable for our study. 
First, since our main interest is to model the predictability of the vowels, we confine the output probability distribution to be over the set of vocalic segments, which is a subset of the phoneme inventory $\mathcal{V}  \subset \boldsymbol{\Phi}$.
Second, we train and evaluate our PLMs to predict the next vowel only in the intra-word positions where we know that the next phoneme is indeed a vowel, given a preceding phoneme context that contains at least one vowel. 
While the output in this modified PLM is over the set $\mathcal{V}$, the word-forms remain sequences in $\mathbf{\Phi}^*$. 
That is, both consonants and vowels could appear in the preceding context. 

Note that we do not employ fixed-length context $n$-gram PLMs in our study since we aim to account for non-local phoneme dependencies within a word-form.
Given that word-forms within a lexicon have arbitrary lengths, restricting the preceding context to a fixed number of phonemes does not enable us to model vowel harmony across variable-length contexts beyond phoneme $n$-grams. 
On the other hand, we do not employ more powerful architectures such as a transformer \citep{vaswani_attention_2017} or a bidirectional LSTM \citep{graves_framewise_2005} on grounds of suitability for the task: (1) the dependencies between vowels are relatively short (the domain of vowel harmony is the phonological word), (2) vowel harmony is a progressive phenomenon (i.e., operates from left to right--unlike its regressive counterpart \textit{umlaut}), and (3) the training sets of the individual languages in our study are likely too small to train a large transformer model. 
Moreover, several prior studies within the information-theoretic approaches to investigate phonological structure have also employed LSTM-based PLMs  \citep[e.g.,][]{10.1162/tacl_a_00296, pimentel-etal-2021-disambiguatory}.

\begin{table*}[!ht]
    \centering
    \resizebox{0.9\textwidth}{!}{%
    \begin{tabular}{l|c|c|c}
    \hline
        \textbf{Language}                   & \multicolumn{3}{c}{\textbf{Harmonic Groups}}                                              \\
        \hline\hline
        \multirow{1}*{Finnish}              & $-\textsc{back}$ \{y, ø, æ\}   & $+\textsc{back}$  \{u, o, \textipa{A}\}  & $\textsc{back}$ neutral  \{e, i\}  \\
        \hline
        \multirow{1}*{Hungarian}            & $-\textsc{back}$ \{y, ø\}   & $+\textsc{back}$  \{u, o, \textipa{6}\} & $\textsc{back}$ neutral    \{e, i\} \\
        \hline
        \multirow{1}*{Manchu}               & $-\textsc{back}$ \{e/\textipa{7}\}  & $+\textsc{back}$  \{\textipa{A}, \textipa{O}\} & $\textsc{back}$ neutral   \{i, u\} \\
        \hline
        \multirow{2}*{Khalkha Mongolian}    & $-\textsc{atr}$  \{e, u, \textopeno\}  & $+\textsc{atr}$  \{a, \textipa{6}, o\}  & $\textsc{atr}$ neutral  \{i\}   \\
                                            & $-\textsc{round}$ \{e, a, i\} & $+\textsc{round}$ \{o\} & $\textsc{round}$ neutral  \{u, \textipa{U}\} \\
        \hline
        \multirow{2}*{Turkish}              & $-\textsc{back}$  \{i, e, y, œ\} & $+\textsc{back}$  \{\textipa{W}, a, u, o\} &   \\
                                            & $-\textsc{round}$  \{i, e, u, o\}& $+\textsc{round}$  \{\textipa{W}, a, y, œ\}&   \\
        \hline
         Arabic, Ainu, Armenian, Basque, Estonian\textdagger  & -- &-- & -- \\
        \hline
    \end{tabular}}
    \caption{Languages from NorthEuraLex used in our sample along with their harmonic groups.
    Khalkha Mongolian has a special type of vowel harmony involving the placement of the tongue
    root: $+\textsc{ATR}$ codes an advanced position of the tongue root in the vocal tract, while $-\textsc{atr}$ encodes an retracted or further back position. Languages in our sample that do not exhibit vowel harmony are marked with the symbol (\textdagger ). }
    \label{tab:langs}
\end{table*}

\subsection{Harmony as Surprisal}

Given that our phoneme-level language model that was trained on a set of word-forms sampled from a natural language lexicon, we can quantify the vowel harmony phenomenon using Shannon's information content, or \textbf{surprisal}. Given a non-initial vocalic position $t$ after a phoneme context $\varphi_{<t}$, vowel surprisal is 

\begin{equation}
    \eta(v, t) = - \text{log}_2 \, p(v \, | \, t, \boldsymbol{\varphi}_{<t})
\end{equation}
which is measured in bits. 
Note that surprisal is maximal when the preceding context tells us nothing about which vowels are more likely to occur. 
That is, if the vowels are sampled from a uniform distribution over the vowel inventory $\mathcal{V}$, then $\eta(v, t) = \text{log}_2 |\mathcal{V}|$ (bits).
Therefore, surprisal in our case is mainly a metric of how ``predictable'' a vowel is in a given context. 
Now consider a set of vowels $\mathcal{H}  \in \mathcal{V}$ that share a phonological feature. 
For a given vowel $v \in \mathcal{H}$,  we refer to the set $\mathcal{H}$ as a harmonic group, while its disharmonic counterpart $\neg\mathcal{H}  \in \mathcal{V} \setminus \mathcal{H}$ as a disharmonic group with respect to the vowel $v$.  
For example, consider the front vowel \text{[i]} in Turkish that has the feature $-\textsc{back}$. 
With respect to \text{[i]}, the front vowels in the Turkish vowel inventory $\mathcal{H} = \{\text{[i]}  ,\text{[e]}, \text{[y]}, \text{[œ]}\}$ make a harmonic group since they all share the feature $-\textsc{back}$, while the rest of the vowels make a disharmonic group $\neg\mathcal{H} =  \{\text{[\textipa{W}]}, \text{[a]}, \text{[u]}, \text{[o]}\}$ since they all lack  the feature $-\textsc{back}$. 
Given a phoneme context that contains at least one vowel $v$ such that $v \in \mathcal{H}$, we compute the surprisal of a harmonic group at position $t$ in a word-form by summing over the vowels in $\mathcal{H}$, i.e.
\begin{equation}
\label{eq:harmonic_surp}
        \eta(\mathcal{H}, t) = - \text{log}_2 \, \sum_{\pi \in \mathcal{H}} p(\pi \, | \, t, \boldsymbol{\varphi}_{<t})
\end{equation}
We refer to the quantity $\eta(\mathcal{H}, t) $ as \textbf{feature surprisal}, since all members of the harmonic group $\mathcal{H}$ share one phonological feature.
Likewise, we compute the surprisal of a disharnomic group by summing over the vowels in $\neg\mathcal{H}$ as
\begin{equation}
\label{eq:disharmonic_surp}
        \eta(\neg\mathcal{H}, t) = - \text{log}_2 \, \sum_{\pi \in \neg\mathcal{H}} p(\pi \, | \, t, \boldsymbol{\varphi}_{<t})
\end{equation}
Assuming that  a PLM has learned the vowel harmony constraints of a language from the training word-forms, we expect the model to predict that vowels in $\mathcal{H}$ are more likely to co-occur in a single word-form.
By implication, we expect the model to ``disfavour'' the occurrence of a  vowel in  $\neg\mathcal{H}$ when observing members of $\mathcal{H}$ in the context.
That is, in a language that exhibits this linguistic phenomenon, word-forms that conform to vowel harmony should be assigned a higher probability than word-forms that do not. 
For example, the Finnish word form [\textipa{s} \textipa{i} \textipa{l} \textipa{m} \textipa{æ} \textipa{s} \textipa{ae}] is expected to be assigned a high probability by our model since the sequence of vowels [\textipa{i}], [\textipa{ae}], [\textipa{ae}] is $-\textsc{back}$ harmonic, and its disharmonic counterpart [s i l m æ s o] is expected to be assigned a lower probability.

Note in equations (\ref{eq:harmonic_surp}) and (\ref{eq:disharmonic_surp}) we compute the surprisal at a single vocalic position in a given word-form. 
To quantify harmonic group surprisal across a set of held-out word-forms $\mathcal{W}$, we compute the quantity 
\begin{equation}
\label{eq:avg_sharmonic_surp}
        \overline{\eta}(\mathcal{H}) = - \frac{1}{|\mathcal{W}|}\sum_{\boldsymbol{w} \in \mathcal{W}} \quad \sum_{t \in \{\tau, \cdots, T\}} \eta(\mathcal{H}, t)
\end{equation}
which is the average feature surprisal. Here, the outer sum $\sum_{\boldsymbol{w} \in \mathcal{W}} $ iterates over all word-forms in $\mathcal{W}$, while the inner sum $\sum_{t \in \{\tau, \cdots, T\}}$ iterates over non-initial vocalic positions within the word-form $\boldsymbol{w}$. 
The feature surprisal of a disharmonic group $\overline{\eta}(\neg\mathcal{H})$ is computed in the same way as in equation (\ref{eq:avg_sharmonic_surp}) but summing over the term $\eta(\neg\mathcal{H}, t)$ instead. 
Finally, we quantify the strength of a vowel harmony constraint in a language as the difference of feature surprisal of the harmonic and disharmonic vowels
\begin{equation}
    \Delta_\eta = \overline{\eta}(\mathcal{H}) - \overline{\eta}(\neg\mathcal{H})
\end{equation}
If feature surprisal in harmonic phoneme sequences is lower than feature surprisal in disharmonic phoneme sequences, $\Delta_\eta$ is negative, indicating that harmonic sequences are assigned higher probability.  
It is worth pointing out that our grouping of the vowels into harmonic groups is only used to obtain feature surprisal values from the model after it has been trained.  
That is, our PLMs for all languages in our study are trained without an explicit signal that informs the model about the features of the vowels.

\section{Experimental Data and Setup}
\label{experi}

\subsection{Data} 

Previous research has made use of large corpora of inflected word-forms \citep{goldsmith_information_2012} or running text \citep{Mayer2010Visua-12872} to infer vowel harmony patterns. 
This is mainly because vowel harmony constraints often surface in inflectional suffixes, especially in highly agglutinating languages such as Finnish, Hungarian or Turkish. 
Though this approach is not in itself problematic, it relies on data that may not exist for the majority of the world's languages. 
It is also not applicable for languages that have a different grammatical structure, for example, reduced or fusional morphology. 
On the other hand, if a language has vowel harmony as a phonologically conditioned rather than a purely grammatical phenomenon, the relevant vowel harmony patterns should also be recoverable from lexical data with little or no inflection at all. 

\begin{table*}[!ht]
    \centering
    \begin{tabular}{lcccc}
        \hline
          & Maximum & Minimum & Average & Median \\
        \hline\hline
        Phoneme inventory size & 72 (Skolt Sami) & 23 (Turkish) & 38.9 & 37\\
        Number of word-forms & 1513 (Manchu) & 677 (Italian) & 1136.6 & 1142 \\
        \hline
    \end{tabular}
    \caption{Inventory sizes and word list lengths in the data sampled from NorthEuraLex.}
    \label{tab:nelex_families}
\end{table*}

We use parts of the NorthEuraLex database (\url{http://www.northeuralex.org/}, \citealt{dellert_northeuralex_2020}) as experimental data to train our phoneme language models and quantify the effect of vowel harmony in languages that are known to exhibit this linguistic phenomenon. 
NorthEuraLex offers a large multilingual word list consisting of 1005 concepts translated into 107 language varieties from North Eurasia with translations provided in a unified transcription following the International Phonetic Alphabet (IPA). 
Moreover, NorthEuraLex contains a larger number of diverse language varieties from various language families that are known to exhibit vowel harmony, as well as language varieties that are known to lack the phenomenon.

As there is no clear definition of what constitutes vowel harmony in languages, and linguistic resources such as the World Atlas of Language Structures \citep{dryer_wals_2014_new} do not provide this information, we concentrate on a subset of 10 language varieties from NorthEuraLex, with five varieties traditionally known to exhibit vowel harmony, and five known to not exhibit the phenomenon. When selecting the languages, we tried to obtain a rather diverse sample of languages from different language families.
Table \ref{tab:langs} gives an overview over the languages and their active harmony processes (where present).
 
The NorthEuraLex data is available in the form of
Cross-Linguistic Data Formats (CLDF \url{https://cldf.clld.org}, \citealt{Forkel2018a}), following the 
recommendations underlying Lexibank \citep{list_lexibank_2022}, a
large collection of lexical word lists (\url{https://github.com/lexibank/northeuralex}).
A core feature of CLDF is the integration of
\emph{reference catalogs}. Reference catalogs are metadata collections that offer basic information
on major linguistic constructs, such as languages (Glottolog, \url{https://glottolog.org},
\citealt{hammarstrom_glottologglottolog_2022}) or concepts (Concepticon, \url{https://concepticon.clld.org},
\citealt{Concepticon}). In addition to offering word lists standardized with respect to language
names and concept elicitation glosses, Lexibank offers standardized
phonetic transcriptions as specified by Cross-Linguistic Transcription Systems (CLTS, \url{https://clts.clld.org}, \citealt{CLTS}), a reference catalog that offers a transcription system
that conforms to the IPA but resolves ambiguities encountered in the
original IPA specification \citep{Anderson2018}. 
 
Since NorthEuraLex is available in CLDF, this means that we have direct access to standardized phonetic
transcriptions segmented into individual sounds in each word form along with an underlying set of
distinctive features provided by CLTS.
The resulting data set provides on average 1136 unique word-forms per language (with several
concepts having two or more word-forms as translational equivalents), with larger differences between individual languages. We decided against downsampling word lists to a common size due to the already small number of samples. The word list sizes range from 971 (Ainu) to 1513 (Manchu).

\subsection{Preprocessing}

For each of the languages, identical word-forms are
collapsed to a single item, such that each sequence of phonemes is presented only once to the model.
In addition, word-forms which are a substring of another word form are also ignored. Thus,
if the word list of a language contains the sequences
\{~[\textipa{s}~\textipa{i}~\textipa{l}~\textipa{m}~\textipa{\ae}],
[\textipa{s}~\textipa{i}~\textipa{l}~\textipa{m}~\textipa{\ae}], 
[\textipa{s}~\textipa{i}~\textipa{l}~\textipa{m}~\textipa{æ}~\textipa{s:}~\textipa{æ}], 
[\textipa{s}~\textipa{i}~\textipa{l}~\textipa{m}~\textipa{\ae}~\textipa{d}~\textipa{\ae}]~\},
only the latter two sequences are kept: 
\{
[\textipa{s}~\textipa{i}~\textipa{l}~\textipa{m}~\textipa{\ae}~\textipa{s:}~\textipa{\ae}], 
[\textipa{s}~\textipa{i}~\textipa{l}~\textipa{m}~\textipa{\ae}~\textipa{d}~\textipa{\ae}]
\}. 
This procedure ensures that only
unique sequences are presented to the model, and that train and test splits do not contain identical forms, which might otherwise lead to unjustified higher weights for sound sequences recurring across the vocabulary of individual language varieties.  

\subsection{Training}
For each language, we randomly split the data into 60\%, 10\% and 30\% subsets for train, validation and test splits respectively. 
The models were trained with the
Adam optimizer \citep{KingmaB14} on the task of minimizing the cross entropy of the predicted
distribution and the true probability distributions over the vowel inventory.  This is equivalent to
minimizing the negative log-likelihood of the true phoneme at each position.  25\% of the inputs
were randomly replaced by a mask token to prevent overfitting on the relatively small sample. Note
that the output probability distribution of the model is restricted to the vowel inventory of the
language plus the end-of-sequence token, since only the vowel positions are of interest for the
analysis.

A separate model was trained for each language in our subset of 10 languages from NorthEuraLex. 
The same hyperparameters were
used for training as in \citet{pimentel_disambiguatory_2021}, with batch size reduced to 32 since
NorthEuraLex wordlists are considerably smaller than the datasets used in that paper. Table
\ref{tab:hyperparams} in Appendix \ref{sec:appendix_hyp} shows the exact configuration of the hyperparameters. 
After each epoch the
models were evaluated on a validation set, and all models were trained until validation loss
converged. Training the models on unique sequences derived from word lists ensures that the model
sees each sequence only once per epoch, and minimizes overlaps between train, test and validation
set. 

\subsection{Significance Tests}

As the expected behavior of vowel harmony languages is that the vowels are not evenly distributed over their words, average feature surprisal is likely to not be normally distributed. 
The Shapiro-Wilk test \citep{shapiro_analysis_1965} was used to check whether the surprisal values for every comparison. For every pairing of conditions at least one of them was not normally distributed with $p < 0.01$.
Thus, the Wilcoxon signed-rank test was conducted to test the significance of a paired contrast (as in the example above). Effect size was calculated as the rank-biserial coefficient using the common language effect size $f = \frac{U}{n_1\cdot n_2}$ as $r = f - (1 - f)$, with $U$ being the test statistic and $n_1 \cdot n_2$ being the number of possible comparisons between two conditions. For an unpaired contrast (e.g. the contrast between average feature surprisal for $+\textsc{round}$ after a $-\textsc{round}$ vowel and average feature surprisal for $+\textsc{back}$ after a $-\textsc{back}$ vowel) a Mann-Whitney U-test was conducted, with effect size calculated as the rank-biserial coefficient using the $T$ statistic 
and the sum of ranks $S$ as $r = \frac{T}{S}$. All significance tests were conducted using the SciPy Python package \citep{virtanen_scipy_2020}.

\subsection{Implementation}
The methods described here are implemented in Python. The PyTorch library \cite{NEURIPS2019_9015} is used to train and evaluate our neural models.
CLDF data are accessed with the help of 
 CL Toolkit (\url{https://pypi.org/project/cltoolkit},
\citealt{list_cl_2021_new}), a Python package that provides convenient access to lexical word lists in
CLDF.

\section{Experimental Results}
\label{results}
\begin{figure*}
    \centering
    \includegraphics[width=\textwidth]{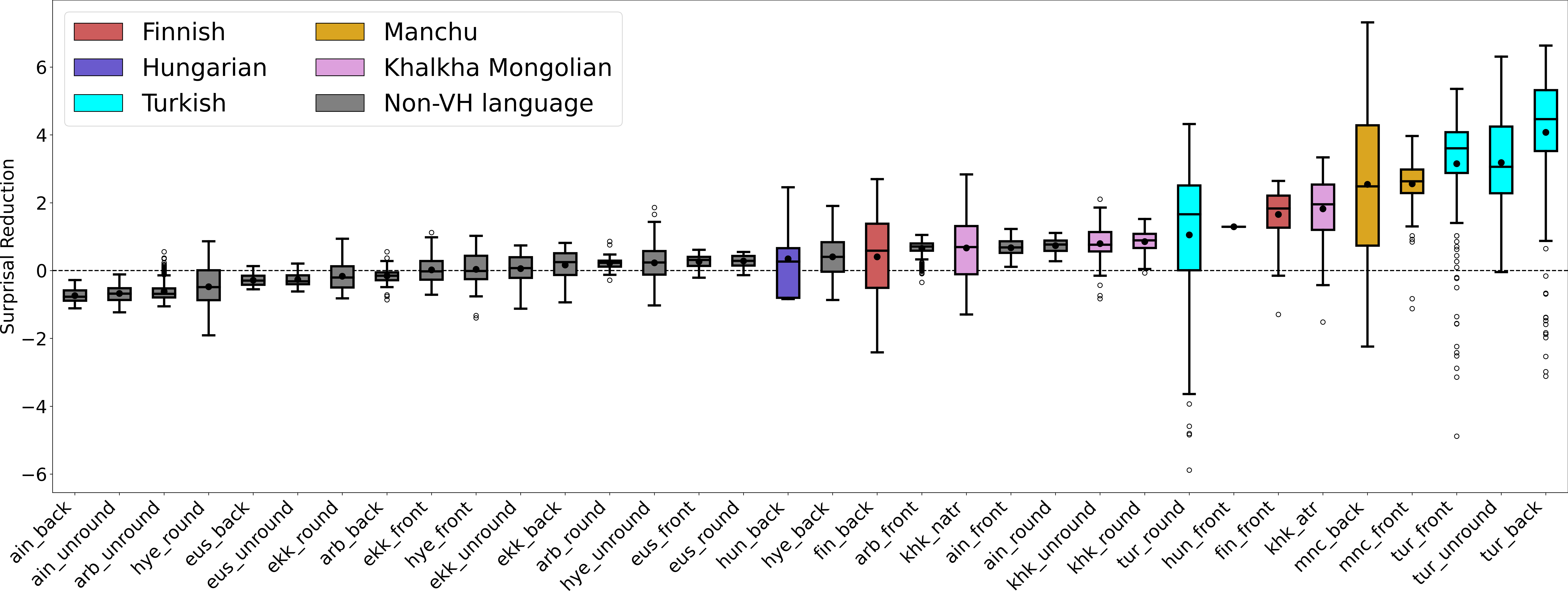}
    \caption{Surprisal reduction for the 10 varieties from NorthEuraLex.  Best viewed in color.}
    \label{fig:surprisal_reduction}
\end{figure*}

\subsection{Feature Surprisal}

All vowel harmony languages show significant differences in feature surprisal between harmonic and
disharmonic conditions with negative $\Delta_\eta$; individual results can be retrieved from the result
tables 6-10 in Appendix \ref{sec:result_tables}. 
Feature surprisal in the $+\textsc{back}$
disharmonic condition was found to be higher than feature surprisal in the $-\textsc{back}$
disharmonic condition for Finnish ($\Delta_\eta = -0.2148,\ p < 0.01$), Hungarian ($\Delta_\eta = -1.0806,\ 
p < 0.01$) and Turkish ($\Delta_\eta = -0.8602,\ p < 0.01$), which confirms the findings of \citet{goldsmith_vowel_1985}.
Note that if the $+\textsc{back}$ and $-\textsc{back}$ harmony were equally strong, one would expect no difference in surprisal if the harmony is violated.
Three out of four languages with $\pm\textsc{back}$ harmony show this
tendency, indicating that the relative strength of $+\textsc{back}$ harmony over $-\textsc{back}$
harmony is the usual case rather than an exception. A possible explanation for this difference in
strength is the existence of neutral vowels, with 3 of the 4 $\pm\textsc{back}$ harmony languages in
our sample having at least one neutral vowel, and Turkish, the only language without neutral
vowels, also showing the largest difference between the two disharmonic conditions . The
probabilities of the neutral vowels are not included in the feature surprisal calculation, causing
feature surprisal to be higher in the $+\textsc{back}$ disharmonic condition while lowering feature
surprisal in the $-\textsc{back}$ disharmonic condition. For Hungarian feature surprisal was lowest
in the neutral harmonic condition, meaning that neutral vowels are most likely to occur after
another neutral vowel. Even though Hungarian neutral vowels trigger $-\textsc{back}$ harmony, the
low number of forms containing both $-\textsc{back}$ vowels and neutral vowels makes it difficult
for the neural language model to learn the pattern, leading to the highest feature surprisal
occurring in the harmonic condition (i.e. for the $-\textsc{back}$ feature).
Figure \ref{fig:surprisal_reduction} gives an overview of the relative strength of vowel harmony for all languages and harmonic features in 
the sample used in this study. For this figure the sign of $\Delta_\eta$ was reversed in order to quantify the reduction of feature surprisal in the harmonic sequences as compared to the disharmonic  sequences for each combination of feature and language. The boxplots of languages without vowel harmony are located towards the left of the plot with small differences between harmonic and disharmonic sequences, with some vowel harmony languages showing similar, yet still positive surprisal reduction (e.g. Finnish $+\textsc{back}$ vowels, Hungarian $+\textsc{back}$ vowels)

\subsection{The Case of Turkish}
\begin{figure}[!b]
    \centering
    \includegraphics[width=0.48\textwidth]{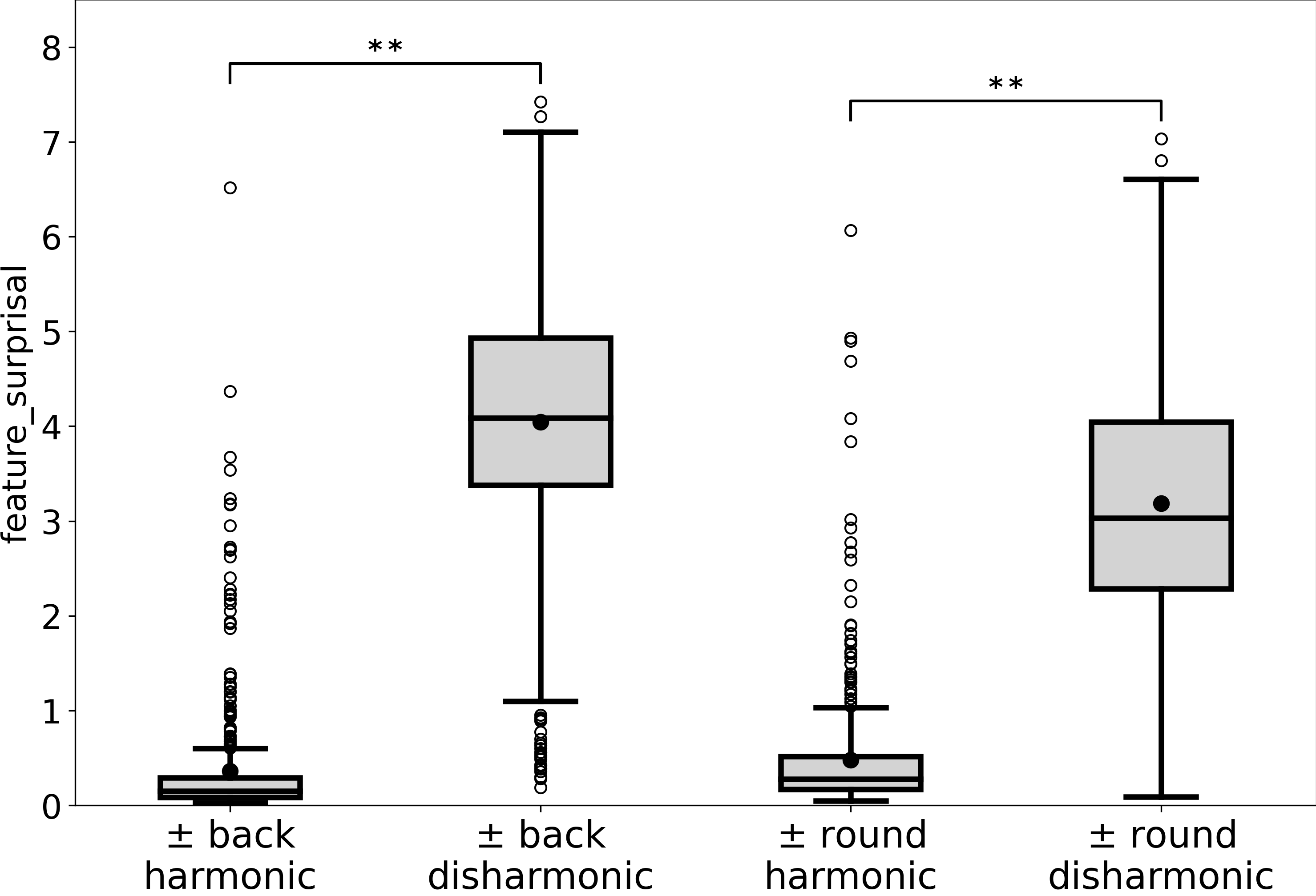}
    \caption{Feature surprisal for Turkish back harmonic/disharmonic sequences (left) and round harmonic/disharmonic sequences (right). The difference between harmonic and disharmonic conditions is significant with $p < 0.01$ in both cases.\textbf{**}: $p < 0.01$, \textbf{*}: $p < 0.05$, \textbf{ns}: $p > 0.05$ }
    \label{fig:res_tur}
\end{figure}

For Turkish the difference in feature surprisal between harmonic and disharmonic conditions was
large. Figure \ref{fig:res_tur} shows that for both the $\pm\textsc{back}$ and $\pm\textsc{round}$
conditions, the disharmonic condition displays a much higher surprisal value as compared to the
harmonic condition ($\Delta_\eta = -3.6816$, $p < 0.01$ and $\Delta_\eta = -2.7061$, $p < 0.01$
respectively). A small but significant bias towards $+\textsc{back}$ harmony was detected ($\Delta_\eta
= -0.8602$, $p < 0.01$). There is one obvious reason for the relative strength of $\pm\textsc{back}$
harmony over $\pm\textsc{round}$, namely the parasitic nature of $\pm\textsc{round}$ harmony in
Turkish: while all morphemes have different forms for $\pm\textsc{back}$, allowing for
$\pm\textsc{round}$ disharmony, only a subset also has separate forms for $\pm\textsc{round}$ (Tab. \ref{tab:vh_turkish}). Thus, there are more instances of $\pm\textsc{back}$ harmony to be
observed by the model, and this is expected to result in higher surprisal values for the
$\pm\textsc{back}$ disharmonic conditions.

After $\pm\textsc{round}$ vowels feature surprisal was also much higher in the disharmonic
conditions, with feature surprisal in the round disharmonic condition being higher than in the
unrounded disharmonic condition ($\Delta_\eta = -1.5827$, $p < 0.01$). 
In other words, $+\textsc{round}$ harmony seems to be stronger than $-\textsc{round}$ harmony in Turkish.
When combining the disharmonic
conditions within a harmonic feature and comparing them to the disharmonic conditions in the other
harmonic feature, the combined back disharmonic condition (both front disharmonic and back
disharmonic) yields slightly higher feature surprisal than the combined rounded disharmonic
condition ($\Delta_\eta = 0.8555$, $p < 0.01$); see Table \ref{tab:res_tur_tests} in the appendix.
This is in line with earlier research \citep{baker_two_2009} that found a bias towards
$\pm\textsc{back}$ harmony over $\pm\textsc{round}$ harmony. This is also the expected result when
taking into account that many suffixes do not have $+\textsc{round}$ forms and therefore introduce
noise to the data.

\subsection{Neutral Vowels}

Learning vowel dependencies across neutral vowels turned out to be difficult: For Manchu and Khalkha
Mongolian the number of test items in this category was so low that no meaningful result could be
produced. This is again caused by the nature of the data which consists of lemma forms. For Finnish
and Hungarian the number of items was sufficient to conduct the appropriate significance tests, but the numbers are still small (102 and 63 respectively). The neural language model did not
learn the association of neutral vowels with  $-\textsc{back}$ as assumed for Finnish and Hungarian,
with significant $\Delta_\eta > 0 $ between the neutral harmonic and neutral disharmonic condition only
for Khalkha Mongolian and $\pm\textsc{atr}$ sequences. In Hungarian, neutral vowels are most likely to occur after other neutral vowels, but this is not the case for Finnish,
Manchu and Khalkha Mongolian. On the other hand, Turkish as the only language in the sample without
neutral vowels showed the largest difference between harmonic and disharmonic conditions for both
$\pm\textsc{back}$ and $\pm\textsc{round}$ (see App. \ref{sec:result_tables} for results).

It may be noted that Turkish, the language with the strongest vowel harmony effect in terms of $\Delta_\eta$, has no neutral vowels both for $\pm\textsc{back}$ and $\pm\textsc{round}$ harmony. This could have facilitated the generalization on the $\pm\textsc{back}$ and $\pm\textsc{round}$ harmony patterns for the neural language model, at least proving that the neural language model does indeed assign higher surprisal to disharmonic sequences, since there the harmony system is symmetrical and the number of vowels is the same for each feature.

\section{Discussion and Conclusion}
\label{conclu}
Prior work in the (computational) linguistics community has adopted information theory as a framework for the study of human language structure across different linguistic levels including phonology \citep[e.g.,][]{10.1162/tacl_a_00296, pimentel-etal-2021-surprisal}, morphology \citep[e.g.,][]{rathi-etal-2021-information, wu-etal-2019-morphological}, and syntax \citep[e.g.,][]{Hahn2018AnIE, futrell-etal-2015-quantifying}. 
Following the same spirit, we have introduced an information-theoretic metric to quantify vowel harmony based on feature surprisal. 
Our experiments have demonstrated that feature surprisal is a good indicator of whether a certain feature participates in vowel harmony patterns in a language, producing significant differences between harmonic and disharmonic conditions for most harmonic features in five vowel harmony languages. 
The effect was found on a very small sample of lemma forms with little to no morphological information,
showing that large amounts of inflectional data are not necessary to identify some, but not all
vowel harmony constraints.  When calculated for $\pm\textsc{back}$ and $\pm\textsc{round}$ features
for five non-vowel harmony languages, the difference in surprisal was close to zero, meaning the neural language
model did not detect any preference for harmony constraints in the languages evaluated.

We showed that neural language models can capture non-local harmony constraints over neutral
vowels, which is not possible with count-based methods as employed by
\citet{Mayer2010Visua-12872} or bigram models as in \cite{goldsmith_information_2012}. Here the
resolution of the analysis is more fine-grained with respect to the features underlying the harmonic
groups. The advantage of the modeling approach presented here over both count-based and
probabilistic models is that it can be used with a small dataset (word lists of about 1000 word-forms, of
which ca. 300 are in the test set as the basis of the actual analysis).

The analysis presented could be extended to other types of phonological constraints, since neural
language models in
theory are able to learn all types of dependencies over sequences of arbitrary length. However,
analysing Finnish, Hungarian, Manchu and Khalkha Mongolian required prior knowledge about harmonic
vowels and the split of vowels into harmonic groups, either because the groups are not defined by
the value of a feature as is the case for languages with neutral vowels, or because the feature
representation in our standardized data itself might
not describe a sound with the feature that is assumed to participate in vowel harmony. 

If it is not known which vowels participate in vowel harmony, it seems best to use information on distinctive features in the data in order to find out which effects can be observed. 
However, if the vowel harmony patterns are as complex as in
Khalkha Mongolian, the approach presented here would probably find its limits in corpus size. 
Identifying the approximate number of distinct word-forms needed to
infer vowel harmony systems of individual language varieties (similar to previous studies inferring
the number of words needed to get
an approximate account of phoneme numbers,
\citealt{dockum_rikker_swadesh_2019}) would be an interesting topic for future analysis.

\section*{Limitations}

The limiting factor in the analysis of Hungarian and Khalkha Mongolian was the low number of items with more than two vowels in the test data. Although this was less of a problem in the other three languages (Finnish, Turkish and Manchu all have 400+ items with three or more vowels), this is likely the case for many of the languages in NorthEuraLex. 
Figure \ref{fig:num_vowels_nelex} in Appendix \ref{sec:appendix} shows that many languages have an even lower number of items with more than three vowels than Finnish and Khalkha Mongolian. 
Given a train-valid-test split of 60\%-10\%-30\%, the number of items available to the analysis of long-range dependencies (including, but not restricted to, the operation of vowel harmony across neutral vowels) will be very low for these languages. This is an inherent property of the data, and could only be amended by using larger word lists or a larger corpora that are not restricted to lemma forms.

\section*{Ethics Statement}
The authors foresee no ethical concerns about the work presented in the paper.

\section*{Supplementary Material}

The supplementary material accompanying this study was archived with Zenodo (\url{https://doi.org/10.5281/zenodo.7782090}). It contains all data and code needed to replicate this study, along with extensive instructions.

\section*{Acknowledgements}
We thank the anonymous reviewers for their constructive comments. This research was supported by the Deutsche Forschungsgemeinschaft (DFG, German Research Foundation), Project ID 232722074 -- SFB 1102 (Julius Steuer, Badr Abdullah), by the Max Planck Society Research Grant \textit{CALC³} (JML, \url{https://digling.org/calc/}), and the ERC Consolidator Grant \textit{ProduSemy} (JML, Grant No. 101044282, see \url{https://doi.org/10.3030/101044282}). Views and opinions expressed are however those of the authors only and do not necessarily reflect those of the European Union or the European Research Council Executive Agency (nor any other funding agencies involved). Neither the European Union nor the granting authority can be held responsible for them.

\bibliography{references,anthology}

\begin{thebibliography}{35}
\expandafter\ifx\csname natexlab\endcsname\relax\def\natexlab#1{#1}\fi

\bibitem[{Anderson et~al.(2018)Anderson, Tresoldi, Chacon, Fehn, Walworth,
  Forkel, and List}]{Anderson2018}
Cormac Anderson, Tiago Tresoldi, Thiago~Costa Chacon, Anne-Maria Fehn, Mary
  Walworth, Robert Forkel, and Johann-Mattis List. 2018.
\newblock \href {https://doi.org/10.2478/yplm-2018-0002} {{A}
  {c}ross-{l}inguistic {d}atabase of {p}honetic {t}ranscription {s}ystems}.
\newblock \emph{Yearbook of the Poznań Linguistic Meeting}, 4(1):21--53.

\bibitem[{Anderson(1980)}]{anderson_problems_1980_new}
Stephen~R. Anderson. 1980.
\newblock \href {https://doi.org/10.1075/slcs.6.02and} {Problems and
  perspectives in the description of vowel harmony}.
\newblock In Robert~M. Vago, editor, \emph{Issues in Vowel Harmony}, volume~6,
  pages 1--48. John Benjamins Publishing Company, Amsterdam.

\bibitem[{Baker(2009)}]{baker_two_2009}
Adam~C. Baker. 2009.
\newblock \href
  {http://www.cs.uchicago.edu/research/publications/techreports/TR2009-03} {Two
  {statistical} {approaches} to {finding} {vowel} {harmony}}.
\newblock Technical report, University of Chicago.

\bibitem[{Crowley(2014)}]{crowley_bislama_2017}
Terry Crowley. 2014.
\newblock \href {https://doi.org/10.1515/9780824850074} {\emph{Bislama
  {reference} {grammar}}}.
\newblock University of Hawaii Press.

\bibitem[{Dellert et~al.(2020)Dellert, Daneyko, Münch, Ladygina, Buch,
  Clarius, Grigorjew, Balabel, Boga, Baysarova, Mühlenbernd, Wahle, and
  Jäger}]{dellert_northeuralex_2020}
Johannes Dellert, Thora Daneyko, Alla Münch, Alina Ladygina, Armin Buch,
  Natalie Clarius, Ilja Grigorjew, Mohamed Balabel, Hizniye~Isabella Boga,
  Zalina Baysarova, Roland Mühlenbernd, Johannes Wahle, and Gerhard Jäger.
  2020.
\newblock \href {https://doi.org/10.1007/s10579-019-09480-6} {{NorthEuraLex}: a
  wide-coverage lexical database of {Northern} {Eurasia}}.
\newblock \emph{Language Resources and Evaluation}, 54(1):273--301.

\bibitem[{Dockum and Bowern(2019)}]{dockum_rikker_swadesh_2019}
Rikker Dockum and Claire Bowern. 2019.
\newblock \href {http://www.elpublishing.org/docs/1/16/ldd16_02.pdf} {Swadesh
  lists are not long enough: {Drawing} phonological generalizations from
  limited data}.
\newblock \emph{Language Documentation and Description}, 16:35--54.

\bibitem[{Dryer et~al.(2014)Dryer, Haspelmath, and
  Forkel}]{dryer_wals_2014_new}
Matthew Dryer, Martin Haspelmath, and Robert Forkel. 2014.
\newblock \href {https://wals.info} {\emph{{WALS} {Online} [{Dataset, Version
  2014.2}]}}.
\newblock Zenodo, Geneva.

\bibitem[{Forkel et~al.(2018)Forkel, List, Greenhill, Rzymski, Bank, Cysouw,
  Hammarström, Haspelmath, Kaiping, and Gray}]{Forkel2018a}
Robert Forkel, Johann-Mattis List, Simon~J. Greenhill, Christoph Rzymski,
  Sebastian Bank, Michael Cysouw, Harald Hammarström, Martin Haspelmath,
  Gereon~A. Kaiping, and Russell~D. Gray. 2018.
\newblock \href {https://doi.org/10.1038/sdata.2018.205} {{C}ross-{L}inguistic
  {D}ata {F}ormats, advancing data sharing and re-use in comparative
  linguistics}.
\newblock \emph{Scientific Data}, 5(180205):1--10.

\bibitem[{Futrell et~al.(2015)Futrell, Mahowald, and
  Gibson}]{futrell-etal-2015-quantifying}
Richard Futrell, Kyle Mahowald, and Edward Gibson. 2015.
\newblock \href {https://aclanthology.org/W15-2112} {Quantifying word order
  freedom in dependency corpora}.
\newblock In \emph{Proceedings of the Third International Conference on
  Dependency Linguistics (Depling 2015)}, pages 91--100, Uppsala, Sweden.
  Uppsala University, Uppsala, Sweden.

\bibitem[{Goldsmith(1985)}]{goldsmith_vowel_1985}
John Goldsmith. 1985.
\newblock \href {https://doi.org/10.1017/S0952675700000452} {Vowel harmony in
  {Khalkha} {Mongolian}, {Yaka}, {Finnish} and {Hungarian}}.
\newblock \emph{Phonology Yearbook}, 2(1):253--275.

\bibitem[{Goldsmith and Riggle(2012)}]{goldsmith_information_2012}
John Goldsmith and Jason Riggle. 2012.
\newblock \href {https://doi.org/10.1007/s11049-012-9169-1} {Information
  theoretic approaches to phonological structure: the case of {Finnish} vowel
  harmony}.
\newblock \emph{Natural Language \& Linguistic Theory}, 30(3):859--896.

\bibitem[{Graves and Schmidhuber(2005)}]{graves_framewise_2005}
Alex Graves and J{\"u}rgen Schmidhuber. 2005.
\newblock \href {https://doi.org/10.1109/IJCNN.2005.1556215} {Framewise phoneme
  classification with bidirectional {LSTM} networks}.
\newblock In \emph{Proceedings. 2005 {IEEE} {International} {Joint}
  {Conference} on {Neural} {Networks}, 2005.}, volume~4, pages 2047--2052,
  Montreal, Que., Canada. IEEE.

\bibitem[{Hahn et~al.(2018)Hahn, Degen, Goodman, Jurafsky, and
  Futrell}]{Hahn2018AnIE}
Michael Hahn, Judith Degen, Noah Goodman, Daniel Jurafsky, and Richard Futrell.
  2018.
\newblock \href {https://mindmodeling.org/cogsci2018/papers/0339/0339.pdf} {An
  information-theoretic explanation of adjective ordering preferences}.
\newblock In \emph{Proceedings of the 40th Annual Meeting of the Cognitive
  Science Society}, pages 1766--1772, Madison, WI. Cognitive Science Society.

\bibitem[{Hammarström et~al.(2022)Hammarström, Forkel, Haspelmath, and
  Bank}]{hammarstrom_glottologglottolog_2022}
Harald Hammarström, Robert Forkel, Martin Haspelmath, and Sebastian Bank.
  2022.
\newblock \href {https://doi.org/10.5281/ZENODO.7398962} {\emph{{Glottolog}
  {[Dataset, Version 4.7]}}}.
\newblock Zenodo, Geneva.

\bibitem[{Hochreiter and Schmidhuber(1997)}]{hochreiter1997long}
Sepp Hochreiter and J{\"u}rgen Schmidhuber. 1997.
\newblock Long short-term memory.
\newblock \emph{Neural computation}, 9(8):1735--1780.

\bibitem[{Kingma and Ba(2015)}]{KingmaB14}
Diederik~P. Kingma and Jimmy Ba. 2015.
\newblock \href {http://arxiv.org/abs/1412.6980} {Adam: {A} method for
  stochastic optimization}.
\newblock In \emph{3rd International Conference on Learning Representations,
  {ICLR} 2015, San Diego, CA, USA, May 7-9, 2015, Conference Track
  Proceedings}.

\bibitem[{List et~al.(2021)List, Anderson, Tresoldi, and Forkel}]{CLTS}
Johann-Mattis List, Cormac Anderson, Tiago Tresoldi, and Robert Forkel. 2021.
\newblock \href {https://clts.clld.org} {\emph{{C}ross-{L}inguistic
  {T}ranscription {S}ystems {[Dataset, {V}ersion 2.1.0]}}}.
\newblock Max Planck Institute for the Science of Human History, Jena.

\bibitem[{List and Forkel(2021)}]{list_cl_2021_new}
Johann-Mattis List and Robert Forkel. 2021.
\newblock \href {https://pypi.org/project/cltoolkit/} {\emph{{CL} {Toolkit}.
  {A} {Python} {library} for the {processing} of {cross}-{linguistic} {data}
  {[Software package, Version 0.1.1]}}}.
\newblock Max Planck Institute for Evolutionary Anthropology, Leipzig.

\bibitem[{List et~al.(2022{\natexlab{a}})List, Forkel, Greenhill, Rzymski,
  Englisch, and Gray}]{list_lexibank_2022}
Johann-Mattis List, Robert Forkel, Simon~J. Greenhill, Christoph Rzymski,
  Johannes Englisch, and Russell~D. Gray. 2022{\natexlab{a}}.
\newblock \href {https://doi.org/10.1038/s41597-022-01432-0} {Lexibank, a
  public repository of standardized wordlists with computed phonological and
  lexical features}.
\newblock \emph{Scientific Data}, 9(1):316.

\bibitem[{List et~al.(2022{\natexlab{b}})List, Tjuka, Rzymski, Greenhill, and
  Forkel}]{Concepticon}
Johann-Mattis List, Annika Tjuka, Christoph Rzymski, Simon~J. Greenhill, and
  Robert Forkel. 2022{\natexlab{b}}.
\newblock \href {https://concepticon.clld.org} {\emph{{CLLD Concepticon}
  [{Dataset, Version 3.0.0}]}}.
\newblock Max Planck Institute for Evolutionary Anthropology, Leipzig.

\bibitem[{Mayer et~al.(2010)Mayer, Rohrdantz, Butt, Plank, and
  Keim}]{Mayer2010Visua-12872}
Thomas Mayer, Christian Rohrdantz, Miriam Butt, Frans Plank, and Daniel~A.
  Keim. 2010.
\newblock Visualizing vowel harmony.
\newblock \emph{Linguistic issues in language technology}, 4(2):1--33.

\bibitem[{Ohala(1994)}]{ohala_towards_1994}
John~J. Ohala. 1994.
\newblock \href {https://doi.org/10.21437/ICSLP.1994-113} {Towards a universal,
  phonetically-based, theory of vowel harmony}.
\newblock In \emph{3rd {International} {Conference} on {Spoken} {Language}
  {Processing} ({ICSLP} 1994)}, pages 491--494. ISCA.

\bibitem[{Ozburn(2019)}]{ozburn_segment-specific_2019}
Avery Ozburn. 2019.
\newblock \href {https://doi.org/10.3765/amp.v7i0.4494} {A {segment}-specific
  {metric} for {quantifying} {participation} in {harmony}}.
\newblock \emph{Proceedings of the Annual Meetings on Phonology}, 7.

\bibitem[{Paszke et~al.(2019)Paszke, Gross, Massa, Lerer, Bradbury, Chanan,
  Killeen, Lin, Gimelshein, Antiga, Desmaison, Kopf, Yang, DeVito, Raison,
  Tejani, Chilamkurthy, Steiner, Fang, Bai, and Chintala}]{NEURIPS2019_9015}
Adam Paszke, Sam Gross, Francisco Massa, Adam Lerer, James Bradbury, Gregory
  Chanan, Trevor Killeen, Zeming Lin, Natalia Gimelshein, Luca Antiga, Alban
  Desmaison, Andreas Kopf, Edward Yang, Zachary DeVito, Martin Raison, Alykhan
  Tejani, Sasank Chilamkurthy, Benoit Steiner, Lu~Fang, Junjie Bai, and Soumith
  Chintala. 2019.
\newblock \href
  {http://papers.neurips.cc/paper/9015-pytorch-an-imperative-style-high-performance-deep-learning-library.pdf}
  {Pytorch: An imperative style, high-performance deep learning library}.
\newblock In H.~Wallach, H.~Larochelle, A.~Beygelzimer, F.~d\textquotesingle
  Alch\'{e}-Buc, E.~Fox, and R.~Garnett, editors, \emph{Advances in Neural
  Information Processing Systems 32}, pages 8024--8035. Curran Associates, Inc.

\bibitem[{Pimentel et~al.(2021{\natexlab{a}})Pimentel, Cotterell, and
  Roark}]{pimentel-etal-2021-disambiguatory}
Tiago Pimentel, Ryan Cotterell, and Brian Roark. 2021{\natexlab{a}}.
\newblock \href {https://doi.org/10.18653/v1/2021.eacl-main.3} {Disambiguatory
  signals are stronger in word-initial positions}.
\newblock In \emph{Proceedings of the 16th Conference of the European Chapter
  of the Association for Computational Linguistics: Main Volume}, pages 31--41,
  Online. Association for Computational Linguistics.

\bibitem[{Pimentel et~al.(2021{\natexlab{b}})Pimentel, Cotterell, and
  Roark}]{pimentel_disambiguatory_2021}
Tiago Pimentel, Ryan Cotterell, and Brian Roark. 2021{\natexlab{b}}.
\newblock Disambiguatory {signals} are {stronger} in {word}-initial
  {positions}.
\newblock In \emph{Proceedings of the 16th {Conference} of the {European}
  {Chapter} of the {Association} for {Computational} {Linguistics}: {Main}
  {Volume}}, pages 31--41, Online. Association for Computational Linguistics.

\bibitem[{Pimentel et~al.(2021{\natexlab{c}})Pimentel, Meister, Salesky,
  Teufel, Blasi, and Cotterell}]{pimentel-etal-2021-surprisal}
Tiago Pimentel, Clara Meister, Elizabeth Salesky, Simone Teufel, Dami{\'a}n
  Blasi, and Ryan Cotterell. 2021{\natexlab{c}}.
\newblock \href {https://doi.org/10.18653/v1/2021.emnlp-main.73} {A
  surprisal{--}duration trade-off across and within the world{'}s languages}.
\newblock In \emph{Proceedings of the 2021 Conference on Empirical Methods in
  Natural Language Processing}, pages 949--962, Online and Punta Cana,
  Dominican Republic. Association for Computational Linguistics.

\bibitem[{Pimentel et~al.(2020)Pimentel, Roark, and
  Cotterell}]{10.1162/tacl_a_00296}
Tiago Pimentel, Brian Roark, and Ryan Cotterell. 2020.
\newblock \href {https://doi.org/10.1162/tacl_a_00296} {{Phonotactic Complexity
  and Its Trade-offs}}.
\newblock \emph{Transactions of the Association for Computational Linguistics},
  8:1--18.

\bibitem[{Polgárdi(1999)}]{polgardi_vowel_1999}
Krisztina Polgárdi. 1999.
\newblock \href {https://doi.org/10.1515/tlir.1999.16.2.187} {Vowel harmony and
  disharmony in {Turkish}}.
\newblock \emph{The Linguistic Review}, 16(2):187--204.

\bibitem[{Rathi et~al.(2021)Rathi, Hahn, and
  Futrell}]{rathi-etal-2021-information}
Neil Rathi, Michael Hahn, and Richard Futrell. 2021.
\newblock \href {https://doi.org/10.18653/v1/2021.emnlp-main.793} {An
  information-theoretic characterization of morphological fusion}.
\newblock In \emph{Proceedings of the 2021 Conference on Empirical Methods in
  Natural Language Processing}, pages 10115--10120, Online and Punta Cana,
  Dominican Republic. Association for Computational Linguistics.

\bibitem[{Rodd(1997)}]{rodd_recurrent_1997}
Jennifer Rodd. 1997.
\newblock \href {https://aclanthology.org/W97-1012} {Recurrent
  {neural}-{network} {learning} of {phonological} {regularities} in {Turkish}}.
\newblock In \emph{{CoNLL97}: {Computational} {Natural} {Language} {Learning}}.

\bibitem[{Shapiro and Wilk(1965)}]{shapiro_analysis_1965}
Sam~S. Shapiro and Martin~B. Wilk. 1965.
\newblock \href {https://doi.org/10.1093/biomet/52.3-4.591} {An analysis of
  variance test for normality (complete samples)}.
\newblock \emph{Biometrika}, 52(3-4):591--611.

\bibitem[{Vaswani et~al.(2017)Vaswani, Shazeer, Parmar, Uszkoreit, Jones,
  Gomez, Kaiser, and Polosukhin}]{vaswani_attention_2017}
Ashish Vaswani, Noam Shazeer, Niki Parmar, Jakob Uszkoreit, Llion Jones,
  Aidan~N Gomez, Łukasz Kaiser, and Illia Polosukhin. 2017.
\newblock \href
  {https://proceedings.neurips.cc/paper/2017/file/3f5ee243547dee91fbd053c1c4a845aa-Paper.pdf}
  {Attention is {all} you {need}}.
\newblock In \emph{Advances in {Neural} {Information} {Processing} {Systems}},
  volume~30. Curran Associates, Inc.

\bibitem[{Virtanen et~al.(2020)Virtanen, Gommers, Oliphant, Haberland, Reddy,
  Cournapeau, Burovski, Peterson, Weckesser, Bright, van~der Walt, Brett,
  Wilson, Millman, Mayorov, Nelson, Jones, Kern, Larson, Carey, Polat, Feng,
  Moore, VanderPlas, Laxalde, Perktold, Cimrman, Henriksen, Quintero, Harris,
  Archibald, Ribeiro, Pedregosa, van Mulbregt, {SciPy 1.0 Contributors},
  Vijaykumar, Bardelli, Rothberg, Hilboll, Kloeckner, Scopatz, Lee, Rokem,
  Woods, Fulton, Masson, Häggström, Fitzgerald, Nicholson, Hagen, Pasechnik,
  Olivetti, Martin, Wieser, Silva, Lenders, Wilhelm, Young, Price, Ingold,
  Allen, Lee, Audren, Probst, Dietrich, Silterra, Webber, Slavič, Nothman,
  Buchner, Kulick, Schönberger, de~Miranda~Cardoso, Reimer, Harrington,
  Rodríguez, Nunez-Iglesias, Kuczynski, Tritz, Thoma, Newville, Kümmerer,
  Bolingbroke, Tartre, Pak, Smith, Nowaczyk, Shebanov, Pavlyk, Brodtkorb, Lee,
  McGibbon, Feldbauer, Lewis, Tygier, Sievert, Vigna, Peterson, More, Pudlik,
  Oshima, Pingel, Robitaille, Spura, Jones, Cera, Leslie, Zito, Krauss,
  Upadhyay, Halchenko, and Vázquez-Baeza}]{virtanen_scipy_2020}
Pauli Virtanen, Ralf Gommers, Travis~E. Oliphant, Matt Haberland, Tyler Reddy,
  David Cournapeau, Evgeni Burovski, Pearu Peterson, Warren Weckesser, Jonathan
  Bright, Stéfan~J. van~der Walt, Matthew Brett, Joshua Wilson, K.~Jarrod
  Millman, Nikolay Mayorov, Andrew R.~J. Nelson, Eric Jones, Robert Kern, Eric
  Larson, C~J Carey, Ilhan Polat, Yu~Feng, Eric~W. Moore, Jake VanderPlas,
  Denis Laxalde, Josef Perktold, Robert Cimrman, Ian Henriksen, E.~A. Quintero,
  Charles~R. Harris, Anne~M. Archibald, Antônio~H. Ribeiro, Fabian Pedregosa,
  Paul van Mulbregt, {SciPy 1.0 Contributors}, Aditya Vijaykumar,
  Alessandro~Pietro Bardelli, Alex Rothberg, Andreas Hilboll, Andreas
  Kloeckner, Anthony Scopatz, Antony Lee, Ariel Rokem, C.~Nathan Woods, Chad
  Fulton, Charles Masson, Christian Häggström, Clark Fitzgerald, David~A.
  Nicholson, David~R. Hagen, Dmitrii~V. Pasechnik, Emanuele Olivetti, Eric
  Martin, Eric Wieser, Fabrice Silva, Felix Lenders, Florian Wilhelm, G.~Young,
  Gavin~A. Price, Gert-Ludwig Ingold, Gregory~E. Allen, Gregory~R. Lee, Hervé
  Audren, Irvin Probst, Jörg~P. Dietrich, Jacob Silterra, James~T Webber,
  Janko Slavič, Joel Nothman, Johannes Buchner, Johannes Kulick, Johannes~L.
  Schönberger, José~Vinícius de~Miranda~Cardoso, Joscha Reimer, Joseph
  Harrington, Juan Luis~Cano Rodríguez, Juan Nunez-Iglesias, Justin Kuczynski,
  Kevin Tritz, Martin Thoma, Matthew Newville, Matthias Kümmerer, Maximilian
  Bolingbroke, Michael Tartre, Mikhail Pak, Nathaniel~J. Smith, Nikolai
  Nowaczyk, Nikolay Shebanov, Oleksandr Pavlyk, Per~A. Brodtkorb, Perry Lee,
  Robert~T. McGibbon, Roman Feldbauer, Sam Lewis, Sam Tygier, Scott Sievert,
  Sebastiano Vigna, Stefan Peterson, Surhud More, Tadeusz Pudlik, Takuya
  Oshima, Thomas~J. Pingel, Thomas~P. Robitaille, Thomas Spura, Thouis~R.
  Jones, Tim Cera, Tim Leslie, Tiziano Zito, Tom Krauss, Utkarsh Upadhyay,
  Yaroslav~O. Halchenko, and Yoshiki Vázquez-Baeza. 2020.
\newblock \href {https://doi.org/10.1038/s41592-019-0686-2} {{SciPy} 1.0:
  fundamental algorithms for scientific computing in {Python}}.
\newblock \emph{Nature Methods}, 17(3):261--272.

\bibitem[{Wu et~al.(2019)Wu, Cotterell, and
  O{'}Donnell}]{wu-etal-2019-morphological}
Shijie Wu, Ryan Cotterell, and Timothy O{'}Donnell. 2019.
\newblock \href {https://doi.org/10.18653/v1/P19-1505} {Morphological
  irregularity correlates with frequency}.
\newblock In \emph{Proceedings of the 57th Annual Meeting of the Association
  for Computational Linguistics}, pages 5117--5126, Florence, Italy.
  Association for Computational Linguistics.

\end{thebibliography}
\bibliographystyle{acl_natbib}

\newpage

\appendix

\flushbottom \onecolumn \sloppy

\section{LSTM Hyperparameters}
\label{sec:appendix_hyp}

\begin{table*}[!ht]
    \centering
    \begin{tabular}{cc}
         \hline
         \textbf{Hyperparameter} & \textbf{Value}  \\
         \hline\hline
         Embedding Size & 32 \\
         Hidden Size    & 256 \\
         LSTM Layers    & 2 \\
         Dropout        & 0.33 \\
         Batch Size     & 32 \\
         \hline
    \end{tabular}
    \caption{Model and Training Hyperparameters as taken from \cite{pimentel_disambiguatory_2021}}
    \label{tab:hyperparams}
\end{table*}

\section{Abbreviations of Harmonic Features}
\label{sec:appendix}
\begin{table*}[!ht]
    \centering
    \begin{tabular}{c|c|c}
        \hline
         \textbf{Abbreviation} & \multicolumn{2}{c}{\textbf{Feature}} \\
         \hline\hline
         b & back & $+\textsc{back}$\\
         f & front & $-\textsc{back}$\\
         r & round & $+\textsc{round}$  \\
         u & unround & $-\textsc{round}$\\
         atr & advanced tongue root & $+\textsc{atr}$\\
         natr & retracted tongue root & $-\textsc{atr}$\\
         n & neutral & \\
         h & harmonic & \\
         dish & disharmonic & \\
         \hline
    \end{tabular}
    \caption{Explanation of the abbreviations used in the result tables. The condition column refers to the type of harmony tested, with vowel successions abbreviated in the way described in this table. The sequence "f\_n\_f" represents sequences starting with a front/$-\textsc{back}$ vowel, followed by a neutral/$\textsc{back}$ neutral vowel and another front/$\-\textsc{back}$ vowel. If more than one harmonic feature is present (as in Turkish, Manchu and Khalkha Mongolian), the magnitude of the effect on feature surprisal is compared between the two features in the disharmonic condition only (compare row "f\_r/dish" in Table \ref{tab:res_tur_tests}). }
    \label{tab:abbreviations}
\end{table*}

\section{Result Tables}\label{sec:result_tables}

\begin{table*}[!ht]
    \centering
    \caption{P-values, $\Delta_\eta$ and effect size for Finnish feature surprisal}
    \begin{tabular}{cccccc}
        \hline
        \textbf{Condition} & $\Delta_\eta$ & \textbf{Statistic} & \textbf{p-value} & \textbf{Effect Size} & \textbf{Test} \\
        \hline \hline
        f\_f/f\_b & -0.8298 &  71.0 & 2.e-12 & 0.0263 & Wilcoxon \\
        b\_b/b\_f & -0.8469 & 415.0 & 3.8e-17 & 0.0572 & Wilcoxon \\
        n\_f/n\_b & 0.0009 & 4800.0 & 0.1723 & 0.4353 & Wilcoxon \\
        f\_b/b\_f & -0.2148 & 3148.0 & 0.001 & -0.2813 & Mann-Whitney\\
        \hline
        f\_n\_f/f\_n\_b & -0.563 &  59.0 & 7.57e-05 & 0.1052 & Wilcoxon \\
        b\_n\_b/b\_n\_f & -0.6077 & 236.0 & 0.0009 & 0.2183 & Wilcoxon \\
        n\_n\_f/n\_n\_b & -0.1206 & 85.0 & 0.1114 & 0.308 & Wilcoxon \\
        f\_n\_b/b\_n\_f & -0.1188 & 688.0 & 0.4834 & -0.0935 & Mann-Whitney \\
        \hline
    \end{tabular}
    \label{tab:res_fin_tests}
\end{table*}

\begin{table*}[!ht]
    \centering
    \caption{P-values, $\Delta_\eta$ and effect size for Hungarian feature surprisal}
    \begin{tabular}{ccccccc}
        \hline
        \textbf{Condition} & $\Delta_\eta$ & \textbf{Statistic} & \textbf{p-value} & \textbf{Effect Size} & \textbf{Test} \\
        \hline \hline
        f\_f/f\_b & -0.0917 &  270.0 & 0.64 & 0.4538 & Wilcoxon \\
        b\_b/b\_f & -2.1995 & 2.0 & 9.46e-21 & 0.0003 & Wilcoxon \\
        n\_f/n\_b & 0.7951 & 1270.0 & 2.47e-14 & 0.1287 & Wilcoxon \\
        f\_b/b\_f & -1.0806 & 364.0 & 5.36e-13 & -0.8154 & Mann-Whitney \\
        \hline
        f\_n\_f/f\_n\_b & 0.0864 &  27.0 & 1.0 & 0.4909 & Wilcoxon \\
        b\_n\_b/b\_n\_f & -1.6036 & 0.0 & 0.0078 & 0.0 & Wilcoxon \\
        n\_n\_f/n\_n\_b & 0.4453 & 243.0 & 0.0019 & 0.2348 & Wilcoxon \\
        f\_n\_b/b\_n\_f & -0.674 & 24.0 & 0.1728 & -0.4 & Mann-Whitney \\
        \hline
    \end{tabular}
    \label{tab:res_hun_tests}
\end{table*}

\begin{table*}[!ht]
    \centering
    \caption{P-values, $\Delta_\eta$ and effect size for Turkish feature surprisal}
    \begin{tabular}{ccccccc}
        \hline
        \textbf{Condition} & $\Delta_\eta$ & \textbf{Statistic} & \textbf{p-value} & \textbf{Effect Size} & \textbf{Test}\\
        \hline \hline
        f\_f/f\_b & -3.1502 &  429.0 & 1.65e-29 & 0.0244 & Wilcoxon \\
        b\_b/b\_f & -4.0729 & 258.0 & 4.25e-42 & 0.008 & Wilcoxon \\
        f\_b/b\_f & -0.8602 & 14301.0 & 9.15e-13 & -0.3978 & Mann-Whitney \\
        \hline
        r\_r/r\_u & -1.0516 &  1107.0 & 1.8e-06 & 0.2236 & Wilcoxon \\
        u\_u/u\_r & -3.185 & 10.0 & 9.0e-58 & 0.0002 & Wilcoxon \\
        r\_u/u\_r & -1.5827 & 6339.0 & 2.48e-21 & -0.6256 & Mann-Whitney \\
        \hline
        f\_h/dish & -3.6816 &  1348.0 & 4.71e-70 & 0.0138 & Wilcoxon \\
        r\_h/dish & -2.7061 & 3473.0 & 4.5e-64 & 0.0356 & Wilcoxon \\
        f/r\_dish & 0.8555 & 132794.0 & 5.55e-21 & 0.3656 & Mann-Whitney\\
        \hline
    \end{tabular}
    \label{tab:res_tur_tests}
\end{table*}

\begin{table*}[!ht]
    \centering
    \caption{P-values, $\Delta_\eta$ and effect size for Manchu feature surprisal}
    \begin{tabular}{ccccccc}
        \hline
        \textbf{Condition} & $\Delta_\eta$ & \textbf{Statistic} & \textbf{p-value} & \textbf{Effect Size} & \textbf{Test} \\
        \hline \hline
        f\_f/f\_b & -2.5563 &  6.0 & 1.68e-24 & 0.0006 & Wilcoxon \\
        b\_b/b\_f & -3.4993 & 209.0 & 1.16e-20 & 0.0253 & Wilcoxon \\
        n\_f/n\_b & 0.354 & 14803.0 & 0.0086 & 0.4076 & Wilcoxon \\
        f\_b/b\_f & 0.1359 & 9167.0 & 0.6778 & 0.0305 & Mann-Whitney \\
        \hline
        f\_n\_f/f\_n\_b & -1.3331 &  43.0 & 3.58e-05 & 0.0814 & Wilcoxon \\
        b\_n\_b/b\_n\_f & -1.5021 & 259.0 & 1.61e-11 & 0.0743 & Wilcoxon \\
        n\_n\_f/n\_n\_b & 0.1291 & 3941.0 & 0.7673 & 0.4849 & Wilcoxon \\
        f\_n\_b/b\_n\_f & -0.0086 & 1273.0 & 0.7338 & -0.0414 & Mann-Whitney \\
        \hline
    \end{tabular}
    \label{tab:res_mnc_tests}
\end{table*}

\begin{table*}[!ht]
    \centering
    \caption{P-values, $\Delta_\eta$ and effect size for Khalkha Mongolian feature surprisal}
    \begin{tabular}{ccccccc}
        \hline
        \textbf{Condition} & $\Delta_\eta$ & \textbf{Statistic} & \textbf{p-value} & \textbf{Effect Size} & \textbf{Test} \\
        \hline \hline
        atr\_atr/atr\_natr & -1.8211 &  27.0 & 1.55e-13 & 0.0095 & Wilcoxon \\
        natr\_natr/natr\_atr & -0.6621 & 1819.0 & 2.55e-12 & 0.1672 & Wilcoxon \\
        n\_atr/n\_natr & -0.6531 & 91.0 & 0.0185 & 0.2407 & Wilcoxon \\
        atr\_natr/natr\_atr & -1.5526 & 7395.0 & 3.21e-05 & 0.3415 & Mann-Whitney \\
        \hline
        r\_r/r\_u & -1.8211 &  2.0 & 4.37e-07 & 0.0034 & Wilcoxon \\
        u\_u/u\_r & -0.6621 & 2.0 & 8.35e-13 & 0.0009 & Wilcoxon \\
        n\_r/n\_u & -0.6531 & 371.0 & 0.148 & 0.3747 & Wilcoxon \\
        r\_u/u\_r & -1.5526 & 170.0 & 2.64e-12 & -0.8529 & Mann-Whitney \\
        \hline
        atr\_h/dish & -1.0537 &  2337.5 & 1.09e-25 & 0.0944 & Wilcoxon \\
        r\_h/dish & -1.6815 & 6.0 & 2.18e-18 & 0.0011 & Wilcoxon \\
        atr/r\_dish & -0.3697 & 8941.0 & 0.0024 & -0.2103 & Mann-Whitney \\
    \hline
    \end{tabular}
    \label{tab:res_khk_tests}
\end{table*}

\clearpage

\newpage
\section{Vowel Counts in Test Set}\label{sec:appendix_c}
\begin{figure*}[!ht]
    \centering
    \includegraphics[width=0.95\textwidth]{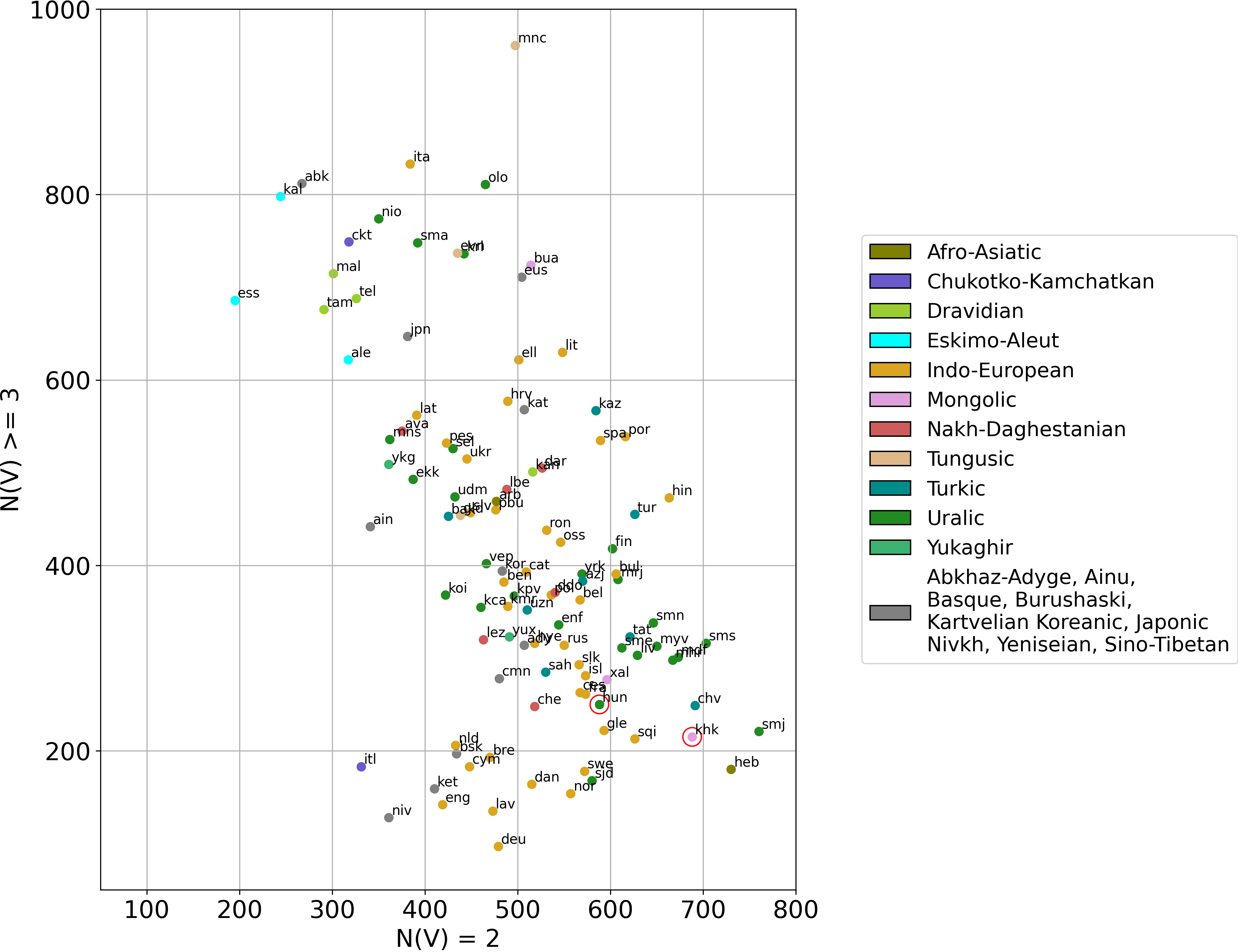}
    \caption{Number of items with 2 vowels (x-axis) and 3 or more vowels (y-axis) in all languages in NorthEuraLex. Hungarian and Khalkha Mongolian in red circles. Languages were coded for language family (see legend) and identified by ISO codes. For a mapping of ISO codes to language see the NorthEuraLex website \url{http://www.northeuralex.org/languages.}}
    \label{fig:num_vowels_nelex}
\end{figure*}

\end{document}